\documentclass[9pt,twocolumn]{article}

\usepackage{graphicx}

\usepackage{amsmath}
\usepackage{amssymb}
\usepackage{nicefrac}       
\usepackage[ruled]{algorithm2e} 

\newcommand{\BB}{B}

\newcommand{\BBhat}{{\hat{ B}}}
\newcommand{\CChat}{{\hat{ C}}}

\newcommand{\XX}{X}
\newcommand{\YY}{Y}
\newcommand{\ZZ}{Z}

\newcommand{\UU}{U}
\newcommand{\VV}{V}
\newcommand{\UUhat}{{\hat{ U}}}
\newcommand{\VVhat}{{\hat{ V}}}

\newcommand{\PP}{P}
\newcommand{\DD}{D}
\newcommand{\QQ}{Q}

\newcommand{\II}{I}

\newcommand{\oneVec}{1}

\newcommand{\RR}{\mathbb{R}}

\newcommand{\diag}{{\rm diag}}
\newcommand{\dMat}{{\rm diagM}}

\DeclareMathOperator*{\argmin}{argmin}

\title{On the Regularization of Autoencoders}
\author{Harald Steck and  Dario Garcia Garcia  \\
  Netflix  \\
  Los Gatos, California\\
  {\tt \{hsteck,dariogg\}@netflix.com}
        }

\begin{document}

\maketitle

\begin{abstract}
  While much work has been devoted to  understanding  the implicit (and explicit) regularization of deep nonlinear networks in the \emph{supervised} setting, this paper focuses on   \emph{unsupervised} learning, i.e.,  autoencoders  are trained with the objective of reproducing the output from the input.
We extend recent results on unconstrained linear models \cite{jin21} and apply them to (1) \emph{nonlinear} autoencoders and
  (2) \emph{constrained} linear autoencoders, obtaining the following two results:
First, we show that the \emph{unsupervised} setting by itself  induces strong additional regularization, i.e., a severe reduction in the model-capacity of the learned autoencoder: we derive that  a \emph{deep nonlinear} autoencoder \emph{cannot} fit the training data more accurately than a \emph{linear} autoencoder does if both models have the same dimensionality in their last hidden layer (and under a few additional assumptions).  Our second contribution is concerned with the  low-rank EDLAE model \cite{steck20b}, which is a linear autoencoder with a constraint on the diagonal of the learned low-rank parameter-matrix for improved generalization: we derive a closed-form approximation to the optimum of its non-convex training-objective, and empirically demonstrate that it is an accurate approximation across \emph{all} model-ranks  in our experiments on three well-known data sets.
\end{abstract}

\section{Introduction}
In recent years, much progress has been made in better understanding various kinds of  implicit  and explicit regularization when training deep nonlinear networks, e.g.,  \cite{neyshabur15b,c_zhang17,c_zhang17b,du18,chaudhari18,nagarajan19,belkin21}. This work has typically been done in the context of \emph{supervised} learning. In this paper, we instead focus on training \emph{autencoders (AE)}, which are used for \emph{unsupervised} learning.
We discuss two kinds of regularization /  restrictions of the model-capacity of the learned AE:
\begin{enumerate}
\item First, we analytically show that a deep nonlinear AE \emph{cannot} fit the data more accurately than a linear AE with the same number of latent dimensions in the last hidden layer does. We argue that  \emph{unsupervised} learning  by itself induces this severe reduction of model-capacity in the deep nonlinear AE, and may hence be viewed as a new kind of implicit regularization.

\item Second, we focus on the Emphasized Denoising Linear Autoencoder (EDLAE) \cite{steck20b}, which is trained with the objective of preventing this linear AE from learning the identity function between its input and output, as to improve generalization. While the resulting training-objective is non-convex, and was optimized iteratively in \cite{steck20b}, we show that there exists a simple yet accurate closed-form approximation.
 
\end{enumerate}

These two contributions are obtained by extending the results on unconstrained linear models in \cite{jin21} to
(1) \emph{deep nonlinear} AEs (first contribution, see Section \ref{sec_nonlin}), and
 (2) \emph{constrained} linear AEs that are prevented from overfitting to the identity function (second contribution, see Section \ref{sec_approx}).
\section{Nonlinear vs Linear Auto-encoders}
\label{sec_nonlin}
In the first part of this paper, we show that the learned \emph{nonlinear} AE cannot fit the data better than a linear AE does if their last hidden layer has the same latent dimensionality.
This is formalized in the following, and various aspects are discussed in the sections thereafter.
\subsection{Training Objective}
While cross-entropy or log-likelihood are popular training-objectives in the deep-learning literature, we here consider the squared error,  as it allows for analytical derivations, see also  \cite{tian21,razin20,belkin21}. 
Also note that minimizing the squared error as a training objective has been related to approximately optimizing ranking metrics   \cite{ramaswamy13,calauzenes20} or classification error  \cite{mai19,muthukumar20,thrampoulidis20,hui21}.
Formally, given training data $X \in \RR^{m\times n}$ regarding $m$ data-points and  $n$ features,
we denote the squared error (SE)  by
\begin{equation}
{\rm SE} = || X - f_\theta(X) ||_F^2  ,
\label{eq_se}
\end{equation}
where the function $f_\theta: \RR^n \rightarrow \RR^n$ represents an arbitrary  AE with model-parameters $\theta$, which is applied row-wise to the data-matrix $X$; and $||\cdot||_F$ is the Frobenius norm.

In addition to the squared error, regularization is typically applied during training, which reduces  model-capacity with the goal of  preventing overfitting. In this first contribution of this paper, we employ a low-dimensional embedding-space as a means of reducing model-capacity--we do not consider any additional explicit regularization here.
 The reason is that  the possibly  high model-capacity of the learned deep nonlinear AE is actually severely reduced in the \emph{unsupervised} setting, even without additional regularization: this is reflected by the fact that 
 a  deep \emph{nonlinear} AE cannot fit the training-data better than a \emph{linear} AE does, as derived below.
The model-fit on the training data can only be further reduced when applying additional regularization. Moreover, as shown in the seminal papers  \cite{neyshabur15b,c_zhang17}, explicit regularization may not be fundamental to understanding the ability of deep nonlinear models to generalize to unseen data.
 \subsection{Autoencoders}
 \label{sec_prop}
In the following, we compare two AEs with each other: a deep-nonlinear AE and a linear AE, which share the following two properties:
\begin{enumerate}
\item the \emph{last} hidden layer has $k$ dimensions, where $k < \min(m,n)$. Note that this (partially) prevents the AE from learning the identity-function. It can severely restrict the model-capacity of the AE for small $k \ll \min(m,n)$. This is the only kind of regularization we consider here.
\item the \emph{output-layer} has a \emph{linear} activation-function. This is a natural choice when using the squared error as training-objective. Note that this choice was also made, for instance,  in \cite{neyshabur15b,liu20}  to facilitate analytical derivations.
\end{enumerate}

We can hence write the deep nonlinear AE in the following form:
\begin{equation}
 f_\theta^{\rm(deep)} (X) = g_{\theta'}(X) \cdot W_L  ,
 \label{eq_deep}
\end{equation}
where the function $g_{\theta'}: \RR^n \rightarrow \RR^k$ is an arbitrary deep nonlinear model parameterized by $\theta'$, which is applied to the data-matrix $X$ in a row-wise  manner;
and $W_L \in \RR^{k\times n}$ is the weight matrix between the last hidden layer (with $k$ dimensions) and the output layer.
Hence, in the nonlinear AE, the  model parameters to be learnt are $\theta = (\theta' ,W_L)$.

The linear AE can be written as 
\begin{equation}
 f_\theta^{\rm(linear)} (X) = X \cdot W_1 \cdot W_2 ,
 \label{eq_lin}
\end{equation}
where $W_1 \in \RR^{n\times k}$ and $W_2 \in \RR^{k\times n}$ are the two weight matrices of rank $k$ to be learnt in the linear AE. 

Having motivated and defined the problem in detail, we are now ready to present the  main result of our first contribution, which will be discussed in the following sections:

{\bf Proposition:} \emph{Consider the squared errors}
\begin{eqnarray}
{\rm SE}^{\rm(deep)} &=& \min_{\theta', W_L}|| X- g_{\theta'}(X) \cdot W_L ||_F^2 \label{eq_se_deep}\\
{\rm SE}^{\rm(linear)} &=& \min_{W_1,W_2}|| X- X \cdot W_1 \cdot W_2 ||_F^2 \label{eq_se_lin} ,
\end{eqnarray}
\emph{of the deep-nonlinear and linear autoencoders on the training data $X\in \RR^{m\times n}$, respectively, 
where the last hidden layer has dimensionality $k < \min(m,n)$ in both autoencoders, i.e., ${\rm rank} (W_L) = {\rm rank} (W_2) = k$. Moreover, note that the output-layer is required to have a linear activation-function in the nonlinear autoencoder in Eq. \ref{eq_se_deep}, while $g_{\theta'}(X)$ can be any deep nonlinear architecture. Then it holds that the deep nonlinear autoencoder \emph{cannot} fit the data better than the linear AE does, i.e., }
\begin{equation}
 {\rm SE}^{\rm(deep)} \ge {\rm SE}^{\rm(linear)} .
 \label{eq_prop}
\end{equation}

{\bf Proof:} This proposition follows from combining two simple mathematical facts: first,
the squared error of the deep non-linear model (which can be viewed as a parametric model) can be bounded by the squared error of a matrix factorization (MF) model (which can be viewed as a nonparametric model), given that the singular value decomposition (SVD) provides the best rank-$k$ approximation (Eckart–Young–Mirsky theorem). Second, the (non-parametric) MF model  can then be rewritten as a (parametric) linear AE, using simple properties of the SVD-solution. This relates the deep-nonlinear AE with the linear AE. The details can be found in the Appendix. $\Box$

\subsection{Discussion}
\label{sec_discuss}

In this section, we discuss several aspects of this seemingly counterintuitive result that a deep-nonlinear AE cannot fit the data better than a linear AE does.

\subsubsection{Unsupervised Learning}

It is crucial for the proof to hold that the \emph{inputs are identical to the targets} when training the model. This \emph{unsupervised}  training-objective is hence responsible for inducing a severe reduction in model-capacity of the deep-nonlinear AE as reflected by its  squared error in the Proposition. This may be viewed as a novel kind of implicit regularization, in addition to the ones discussed in the literature, e.g., \cite{neyshabur15b,c_zhang17,c_zhang17b,du18,chaudhari18,nagarajan19,belkin21} etc.
It is important to note that these results  do \emph{not} apply to deep nonlinear models in general where the input-vector is \emph{different} from the target-vector during training, like in supervised learning.

{\bf Recommender Systems: }
An important  application of unsupervised learning of AEs  is the  'classic' recommendation problem in the literature, where only the user-item interaction-data $X$ are available, which are then randomly split into disjoint training and test data.\footnote{In contrast, a scenario that we do not consider to be a 'classic' recommendation problem, is the prediction of the next item in a sequence of user-item interactions.} The empirical observations in the literature have been puzzling so far, given that  (1) among the deep nonlinear AEs, rather shallow architectures with typically only 1-3 hidden layers were empirically found to obtain the highest ranking accuracy on the test data \cite{sedhain15, liang18, shenbin20,lobel20,khawar20}; and (2) simple linear AEs were recently found to be competitive with deep nonlinear AEs \cite{steck19a, steck20b}.
The derived proposition may provide an explanation for these otherwise counterintuitive  empirical findings. 
While our result only applies to the training-error, the model capacity can only be further reduced by applying additional regularization  to improve generalization / test-error.  Given that deep nonlinear AEs typically have a more complex architecture than linear AEs do, this may provide more 'knobs' for applying various kinds of regularization, e.g., \cite{sedhain15, liang18, shenbin20,lobel20,khawar20}, which may eventually result in improved test-errors, compared to linear AEs. 
The empirical observations  that deep nonlinear AEs achieve slightly better test-errors on some datasets, while linear AEs are better on others, e.g., \cite{steck19a}, suggest that  deep-nonlinear AEs only appear to be extremely flexible models, while in fact their prediction accuracy may be comparable to the one of linear AEs after adding proper regularization, as corroborated by the derived proposition regarding the training error.

\subsubsection{Weighted Loss}
\label{sec_limit}
\emph{Weighted} loss-functions have been widely adopted in recent years. Two scenarios have to be distinguished: (1) a weight applies to a data point (i.e., row in $X$), and (2) a weight applies to a feature of a data-point (i.e.,  entry in $X$). Examples of the former include  off-policy evaluation, reinforcement learning, and causal inference, while an example of the latter is weighted matrix factorization for collaborative filtering, e.g., \cite{hu08,pan08}.

  The Proposition obviously carries over to the first scenario, where the weight applies to  a row in $X$,  as a weight can be viewed as the multiplicity with which a data-point appears in the data. 
In contrast, the Proposition does not immediately carry over to the second scenario, as the proof hinges on the singular value decomposition, which does not allow for a different weight being applied regarding each entry (rather than row) of $X$. 

\subsubsection{Representation Learning}

Despite the limitations in terms of accuracy, a deep-nonlinear AE may still have advantages over a linear AE, namely when it comes to representation learning:
a deep nonlinear AE may be able to learn a \emph{lower}-dimensional  embedding (i.e., with dimensionality less than $k$) in one of the layers before the last hidden layer without a (significant) loss in accuracy. Such an increased dimensionality reduction can be beneficial in the subsequent step when these lower-dimensional embeddings are used as features in a supervised model, resulting in improved classification accuracy of the final supervised model, e.g., \cite{rifai11}.

\subsubsection{Neural Tangent Kernel}
In the regime of the \emph{neural tangent kernel} (NTK), i.e., in the limit of infinite width of the deep nonlinear architecture,   neural networks (with a linear output-layer)  behave like linear models \cite{jacot18,liu20}. Note that this connection to linear models is unrelated to the scenario considered here, as we are concerned with \emph{unsupervised} learning of AEs that have a \emph{limited} rank/width $k<\min(m,n)$.


\section{Approximate  EDLAE}
\label{sec_approx}

In the second contribution of this paper, after a brief review of the EDLAE model \cite{steck20b} in the following section, 
we show that there exists an (approximate) \emph{closed-form} solution for training a low-rank EDLAE model (Section \ref{sec_derivation}).
The derived solution results in Algorithm \ref{algo}, which consists of only four steps (Section \ref{sec_algo}). In Section \ref{sec_exp},  we  observe that the approximation accuracy is very high for all model-ranks $k$ on all three well-known data-sets used in our experiments, compared to the (exact) iterative approach (ADMM) used in \cite{steck20b}.  Hence,
iterative approaches to optimize the non-convex training-loss of EDLAE   may actually not be necessary. 

\subsection{Brief Review of EDLAE}
\label{sec_edlae}

In \cite{steck20b}, it was shown that it is crucial to prevent the AE from learning the identity function between its input and output. This can be achieved by the stochastic training-procedure called \emph{emphasized denoising} \cite{vincent10} (in particular for so-called \emph{full emphasis}), which was shown in \cite{steck20b} to be equivalent to modifying the least-squares objective of a linear low-rank AE as follows:
\begin{eqnarray}
&&\lVert \XX - \XX \cdot\left\{\UU\VV^\top -\dMat\left(\diag(\UU\VV^\top )\right)\right\} \rVert_F^2 \nonumber\\
 &&+ \lVert \Lambda^{\nicefrac{1}{2}}  \cdot\left\{\UU\VV^\top -\dMat\left(\diag(\UU\VV^\top )\right)\right\} \rVert_F^2 ,
\label{eq_trainuv_orig}
\end{eqnarray}
where $\XX \in \RR^{m\times n}$ is the given training data,
and the weight matrices $\UU,\VV \in \RR^{n\times k}$ of rank $k$ are the parameters of the low-rank EDLAE  to be learnt. The L$_2$-norm regularization is controlled by the diagonal matrix
\begin{equation}
 \Lambda = \lambda \cdot\II + \frac{p}{q} \cdot \dMat(\diag(\XX^\top \XX) ),
\label{eq_l2}
\end{equation}
 where $p\in[0,1]$ is the dropout probability in emphasized denoising, and $q=1-p$; '$\dMat$' denotes a diagonal matrix, while '$\diag$' denotes the diagonal of a matrix. We added $\lambda\in \RR$ as an additional regularization parameter (and $\II$ denotes the identity matrix), which provides the same regularization across all features, while the second term (controlled by $p$) is feature-specific.
The key result of the analytic derivation in \cite{steck20b} is that the diagonal $\diag(\UU\VV^\top )$ gets ignored/removed from the matrix $\UU\VV^\top$ when learning its parameters in Eq. \ref{eq_trainuv_orig}.

\subsection{Closed-form Solution}
\label{sec_derivation}
This section outlines the derivation of the closed-form solution that approximates the learned EDLAE model, i.e., the optimum of Eq. \ref{eq_trainuv_orig}. Following \cite{jin21},
we first  rewrite the training objective in Eq. \ref{eq_trainuv_orig} as follows: 
with the definitions
\begin{equation}
 \YY =\left [ \begin{array}{c} \XX \\ 0 \end{array} \right ]\,,
 \,\,\,\,\,\,\,
 \ZZ =\left [ \begin{array}{l} \XX \\ \Lambda^{\nicefrac{1}{2}} \end{array} \right ]\,,
\end{equation}
Eq. \ref{eq_trainuv_orig} is equal to
\begin{equation}
 \lVert \YY - \ZZ \cdot\left\{\UU\VV^\top -\dMat\left(\diag(\UU\VV^\top )\right)\right\} \rVert_F^2 .
\label{eq_trainuv_2}
\end{equation}
Now, introducing the \emph{full-rank} solution $\BBhat \in \RR^{n\times n}$ of the minimization problem in Eq. \ref{eq_trainuv_2}, we obtain
\begin{eqnarray}
 &&\lVert \YY - \ZZ \left\{\UU\VV^\top -\dMat\left(\diag(\UU\VV^\top )\right)\right\} \rVert_F^2 \nonumber\\
 &=&
 \lVert \YY - \ZZ \left\{\BBhat -\dMat\left(\diag(\BBhat )\right)\right\}\nonumber\\
 &&\,\,\,\, +\ZZ \left\{\BBhat -\dMat\left(\diag(\BBhat )\right)\right\}\nonumber\\
 &&\,\,\,\, -\ZZ \left\{\UU\VV^\top -\dMat\left(\diag(\UU\VV^\top )\right)\right\} \rVert_F^2 \nonumber\\
 &=&
 \lVert \YY - \ZZ \left\{\BBhat -\dMat\left(\diag(\BBhat )\right)\right\} \rVert_F^2\nonumber\\
 &&\!\!\!\!\!+\lVert \ZZ \left\{\BBhat -\dMat\left(\diag(\BBhat )\right)\right\}\nonumber\\
 &&\,\,\,\,\, -\ZZ \left\{\UU\VV^\top -\dMat\left(\diag(\UU\VV^\top )\right)\right\} \rVert_F^2 
\label{eq_train_2part}
\end{eqnarray}
Note that the second equality holds for the following two reasons: (1) because $\BBhat$ is the \emph{optimum} of the least-squares problem
$\lVert \YY - \ZZ \{\BB -\dMat(\diag(\BB ))\} \rVert_F^2$,
 the residuals $ \YY - \ZZ \{\BBhat -\dMat(\diag(\BBhat ))\}$ hence have to be orthogonal to the predictions $\ZZ \{\BBhat -\dMat(\diag(\BBhat ))\}$; (2) the low-rank model $\UU\VV^\top$ is obviously nested within the full-rank model $\BB$, in other words, it lives in the linear span of the full-rank model: hence  the difference $ \{\BBhat -\dMat(\diag(\BBhat ))\} - \{\UU\VV^\top -\dMat(\diag(\UU\VV^\top ))\}$ also lives in the linear span of the full rank model. Hence we have that $ \ZZ ( \{\BBhat -\dMat(\diag(\BBhat ))\} - \{\UU\VV^\top -\dMat(\diag(\UU\VV^\top ))\})$ is orthogonal to the residuals $ \YY - \ZZ \{\BBhat -\dMat(\diag(\BBhat ))\}$, and hence the trace of their product vanishes, which eliminates the cross-term in Eq. \ref{eq_train_2part}. This is also corroborated by our experiments, where we computed the squared errors of the learned models in these different ways.

Based on the simplification in Eq. \ref{eq_train_2part},
the low-rank optimization problem can hence be decomposed into  the following two steps:
\begin{enumerate}
 \item Computing the full-rank solution $\BBhat\in \RR^{n\times n}$: its closed-form solution is given by \cite{steck20b}:\footnote{Note that $\BB -\dMat(\diag(\BB ))$ may seem to leave the learned diagonal of $\BB$ unspecified. As outlined in \cite{steck20b}, it makes sense to set the diagonal to zero in the full rank model $\BB$, where all the parameters are independent of each other. Note that this is different from the situation in the \emph{low-rank} model $\UU\VV^\top$, where the diagonal $\diag(\UU\VV^\top )$ is specified by minimizing Eq. \ref{eq_trainuv_2}: the off-diagonal elements of $\UU\VV^\top$ are fit to the data, which in turn determines the diagonal $\diag(\UU\VV^\top )$ due to the low-rank nature of $\UU\VV^\top$. Note that $\diag(\UU\VV^\top )\ne 0$ in general.}
\begin{eqnarray}
 \BBhat &=& \argmin_\BB \lVert \YY - \ZZ \cdot\left\{\BB -\dMat\left(\diag(\BB )\right)\right\} \rVert_F^2 \nonumber\\
 &=& \argmin_{\BB \,\,{\rm s.t.\,\,}\diag(\BB)=0 } \lVert \YY - \ZZ\BB \rVert_F^2 \nonumber\\
 &=& \II -\CChat \cdot \dMat(\oneVec \oslash \diag(\CChat)) ,
 \label{eq_bb}
\end{eqnarray}
where $\oslash$ denotes the element-wise division of vectors, and 
\begin{equation}
\CChat =  \left(\ZZ^\top \ZZ \right)^{-1} = \left(\XX^\top \XX + \Lambda \right)^{-1},
\end{equation}
where  the inverse exists for appropriately chosen L$_2$-norm regularization $\Lambda$.

\item Estimating the low-rank solution $\UUhat,\VVhat\in \RR^{n\times k}$ of\footnote{Note that we used $\diag(\BBhat)=0$, see previous footnote and Eq. \ref{eq_bb}.}
  \begin{equation}
\min_{\UU,\VV} \lVert \ZZ \BBhat
-\ZZ \left\{\UU\VV^\top -\dMat\left(\diag(\UU\VV^\top )\right)\right\} \rVert_F^2.
\label{eq_part2}
  \end{equation}
It is important to realize that the diagonal values of the low-rank model $\UU\VV^\top$ do not matter (as they get cancelled) in this optimization. In other words, only the \emph{off-diagonal} elements of the low-rank model $\UU\VV^\top$ have to be fitted.

In the following, we propose a simple yet accurate approximation to this optimization problem by essentially ignoring the diagonal. To this end,  let us assume that we knew the optimal values of the diagonal $\hat{\beta}:=\diag(\UUhat\VVhat^\top )$. Then the optimization problem becomes
$$
\min_{\UU,\VV} \lVert \ZZ \{\BBhat + \dMat(\hat{\beta})\}
-\ZZ \UU\VV^\top \rVert_F^2,
$$
where $\hat{\beta}$ is added to the zero-diagonal of $\BBhat$, and the low-rank model $\UU\VV^\top$ gets fitted. Let us now
consider the contributions to this squared error from the diagonal vs. off-diagonal elements of $\UU\VV^\top$. The first observation is that the contribution from the diagonal vanishes for the optimal $\UUhat\VVhat^\top$, given that   $\hat{\beta}=\diag(\UUhat\VVhat^\top )$. As we generally do not know $\hat{\beta}$, we assume that we can instead use a 'reasonable' approximation $\beta$ such that the average squared error regarding a diagonal element, i.e., $|\beta - \hat{\beta}|^2 / n $, is of the same order of magnitude as the average squared error regarding an off-diagonal element.
In the case that we increase the model-rank $k$ such that it approaches $n$ (full-rank), we know that the optimal $\hat{\beta}$ approaches 0, see full-rank solution in Eq. \ref{eq_bb}. In this case, $\beta=0$ is hence a reasonable choice. Moreover, for arbitrary model-rank $k$, given that  $\UU\VV^\top$ is learned with the objective of predicting a feature from other (typically similar) features (due to emphasized denoising, see Section \ref{sec_edlae}), one can expect that the diagonal element  $(\UUhat\VVhat^\top)_{i,i}$ regarding a feature $i$ is of the same order of magnitude as the largest element in the $i$-th column   $(\UUhat\VVhat^\top)_{-i,i}$, where $-i$ denotes the set of all indices $j\ne i$ (i.e., the largest value is associated with a feature $j$ that is similar to feature $i$). We can use this heuristic to obtain reasonable values for $\beta$. But also when we simply use $\beta=0$ here, the squared error regarding each element on the diagonal  on average is not orders of magnitude larger, compared to the average squared error regarding an off-diagonal element. This is also corroborated by our experiments, where we tried these heuristics (and a few more), and found that they all  resulted in very small differences in the solutions, often within the confidence intervals of the ranking metrics on the test data. We hence show results for the simple choice $\beta=0$ in the experiments in Section \ref{sec_exp}. 

Now that we have seen that all the squared errors (on and off the diagonal)  are on average of the same order of magnitude,
the second observation is that  there are $n$ diagonal elements, while there are $n^2-n$ off-diagonal elements--hence the diagonal elements are outnumbered by the off-diagonal ones if there is a reasonably large number $n$ of features, i.e., when $n \ll n^2-n$, which clearly holds in many  real-world applications, where the number of features is often in the hundreds, thousands or even millions. Hence, the aggregate squared error of all the off-diagonal elements can be (possibly several) orders of magnitude larger than the aggregate squared error of all the diagonal elements. For this reason, any 'reasonable' choice $\beta$  (like $\beta=0$) suffices for obtaining a very accurate approximation of the optimal $\UUhat\VVhat^\top$ if the number of features $n$ is large, as confirmed by our experiments.

With the choice $\beta=0$, the approximate optimization problem hence becomes
\begin{equation}
\min_{\UU,\VV} \lVert \ZZ \BBhat 
-\ZZ \UU\VV^\top \rVert_F^2,
\label{eq_approx}
\end{equation}
which can easily be solved: let the singular value decomposition (SVD) of $ \ZZ \BBhat $ be given by 
$$
\ZZ \BBhat = \PP\DD\QQ^\top,
$$
where $\DD$ is the diagonal matrix of singular values, while the matrices $\PP$ and $\QQ$ are composed of the left and right singular vectors, respectively. Then, according to the well-known Eckart–Young–Mirsky theorem, the optimal rank-$k$ solution of Eq. \ref{eq_approx} is given by $\PP_k\DD_k\QQ_k^\top$ pertaining to the largest $k$ singular values. Given that \cite{jin21}
$$
\PP_k\DD_k\QQ_k^\top
=\PP\DD\QQ^\top \QQ_k \QQ_k^\top
= \ZZ \BBhat \QQ_k \QQ_k^\top,
$$
and comparing the last expression with Eq. \ref{eq_approx}, we obtain the solution of Eq. \ref{eq_approx}:
\begin{eqnarray}
 \UUhat &=& \BBhat \QQ_k \label{eq_uuhat}\\
 \VVhat &=& \QQ_k \label{eq_vvhat},
\end{eqnarray}
which is  the approximate solution of Eq. \ref{eq_part2}

\end{enumerate}

As to compute $\QQ_k$ in practice, instead of using SVD on $\ZZ\BBhat$, which is an $m \times n$ matrix, it might be more efficient to determine the eigenvectors of the $n\times n$ matrix $\BBhat^\top \ZZ^\top \ZZ \BBhat = \BBhat^\top (\XX^\top \XX+\Lambda) \BBhat$. Note that the latter matrix can also be computed more efficiently:
\begin{equation*}
\BBhat^\top \ZZ^\top \ZZ \BBhat = \ZZ^\top \ZZ - \dMat(\oneVec \oslash \diag(\CChat))\cdot (\II +\BBhat).
\end{equation*}

\subsection{Algorithm}
\label{sec_algo}
The derived closed-form solution that approximates the optimal low-rank EDLAE gives rise to a particularly simple algorithm, as shown in Algorithm \ref{algo}: in the first two lines, the full-rank solution is computed in closed form. In the third line, 'eig$_k$' denotes the function that returns the $k$ eigenvectors pertaining to the $k$ largest eigenvalues, which yields the solution $\VVhat$ (see also Eq. \ref{eq_vvhat}). In the fourth line, the solution $\UUhat$ is computed (see also Eq. \ref{eq_uuhat}).

\begin{algorithm}[b]
\SetAlgoNoLine
\KwIn{data matrix $\XX^\top \XX \in \RR^{n\times n }$, \\
 \hspace{11 mm} L$_2$-norm regularization $\Lambda$, see Eq. \ref{eq_l2},\\
 \hspace{11 mm} rank $k$ of low-rank solution.}
\KwOut{approx. low-rank solution $\UUhat,\VVhat \in \RR^{n\times k }$.}
$\CChat = \left(\XX^\top \XX + \Lambda \right)^{-1}$ \\
$ \BBhat = \II -\CChat \cdot \dMat(\oneVec \oslash \diag(\CChat)) $\\
$\VVhat = {\rm eig}_k ( \BBhat^\top (\XX^\top \XX+\Lambda) \BBhat )$\\
$\UUhat = \BBhat \VVhat$ \\
\caption{Pseudocode for the approximate closed-form solution of Eq. \ref{eq_trainuv_orig}.}
\label{algo}
\end{algorithm}

The computational cost of this closed-form approximation is considerably smaller than the cost of the iterative optimization (ADMM) used in \cite{steck20b} unless the model-rank $k$ is very large (see Table \ref{tab_runtime}). In both approaches, an $n \times n$ matrix has to be inverted,
which has computational complexity of about ${\cal O}(n^{2.376})$  when using the Coppersmith-Winograd algorithm. 
The key difference is that the leading $k$ eigenvectors have to be computed in the proposed approximation, while each step of the iterative ADMM updates involves several matrix multiplications  of size $n \times k$. While computing all eigenvectors of a symmetric matrix of size $n\times n$ has  the same computational complexity as  matrix multiplication or inversion \cite{pan99}, i.e.,  ${\cal O}(n^{2.376})$, computing the top $k$ eigenvectors for small $k$ can have a smaller cost, typically  ${\cal O}(kn^2)$, e.g., using the Lanczos algorithm \cite{lanczos50,lehoucq98}. An empirical comparison of the training-times is shown in Table \ref{tab_runtime}: as we can see, the closed-form approximation is up to three times faster than ADMM (which we ran for 10 epochs here).\footnote{Note that 10 epochs is often the minimum that yields 'acceptable' accuracy in practice \cite{boyd11}.} Only for very large model-ranks $k$, ADMM is faster on the \emph{ML-20M} data set in Table \ref{tab_runtime}.

\begin{table}[t]
\caption{Runtimes  (wall clock time in seconds) of ADMM used in \cite{steck20b} and of the proposed approximation for different model-ranks $k$ on the \emph{ML-20M} dataset, using an AWS instance with 128 GB memory and 16 vCPUs. For simplicity, the times reported for ADMM are based on 10 epochs, even though ADMM may not have  reached the optimal test error.$^{\rm 4}$}
\label{tab_runtime}
\begin{tabular}{lrrrrr}
\hline
rank $k$  & 10    &    500  & 1,000  &    2,000   & 5,000\\
\hline
ADMM       &   519s   & 557s &  597s     & 701s   & 1,040s\\
Algo. 1     &  171s   &    191s    & 220s  & 393s    &   1,496s \\
\hline
speed-up    & $3.0\times$     & $2.9\times$  & $2.7\times$    & $1.8\times$  & $0.7\times$ \\
\hline
\end{tabular}
\end{table}

As an aside, this approach may also be viewed as a simple  example of teacher-student learning for  knowledge distillation, which was introduced in   \cite{bucilua06}: first, the \emph{teacher}  is learned from the given data, which is then followed by learning the \emph{student}   based on the predictions of the teacher. In our case, the key step is to learn the 'correct' \emph{off-diagonal} elements of the full-rank model $\BBhat$ from the given data, which is achieved by constraining its diagonal to zero. In the second step, the low-rank student $\UUhat\VVhat^\top$  then learns these off-diagonal values from the teacher $\BBhat$.

\subsection{Experiments}
\label{sec_exp}
In this section, we empirically assess the approximation-accuracy of the closed-form solution derived above. To this end, we follow the  experimental protocol as in \cite{steck20b} and \cite{liang18}, using their publicly available Python notebooks.\footnote{{\tt https://github.com/hasteck/EDLAE\_NeurIPS2020} and {\tt https://github.com/dawenl/vae\_cf}}

\begin{figure}[h!]
 \centering
 \includegraphics[width=58mm]{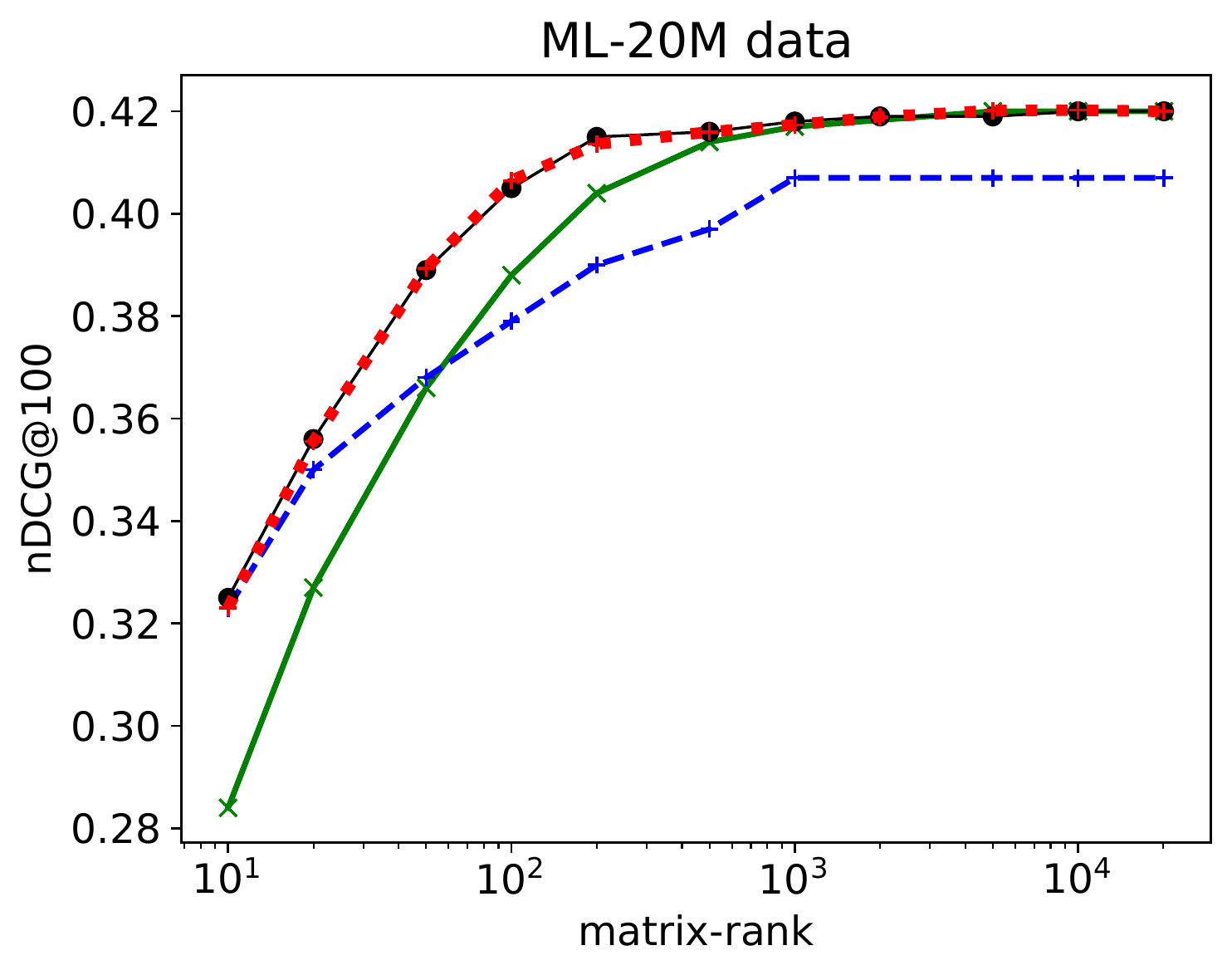}
 \includegraphics[width=58mm]{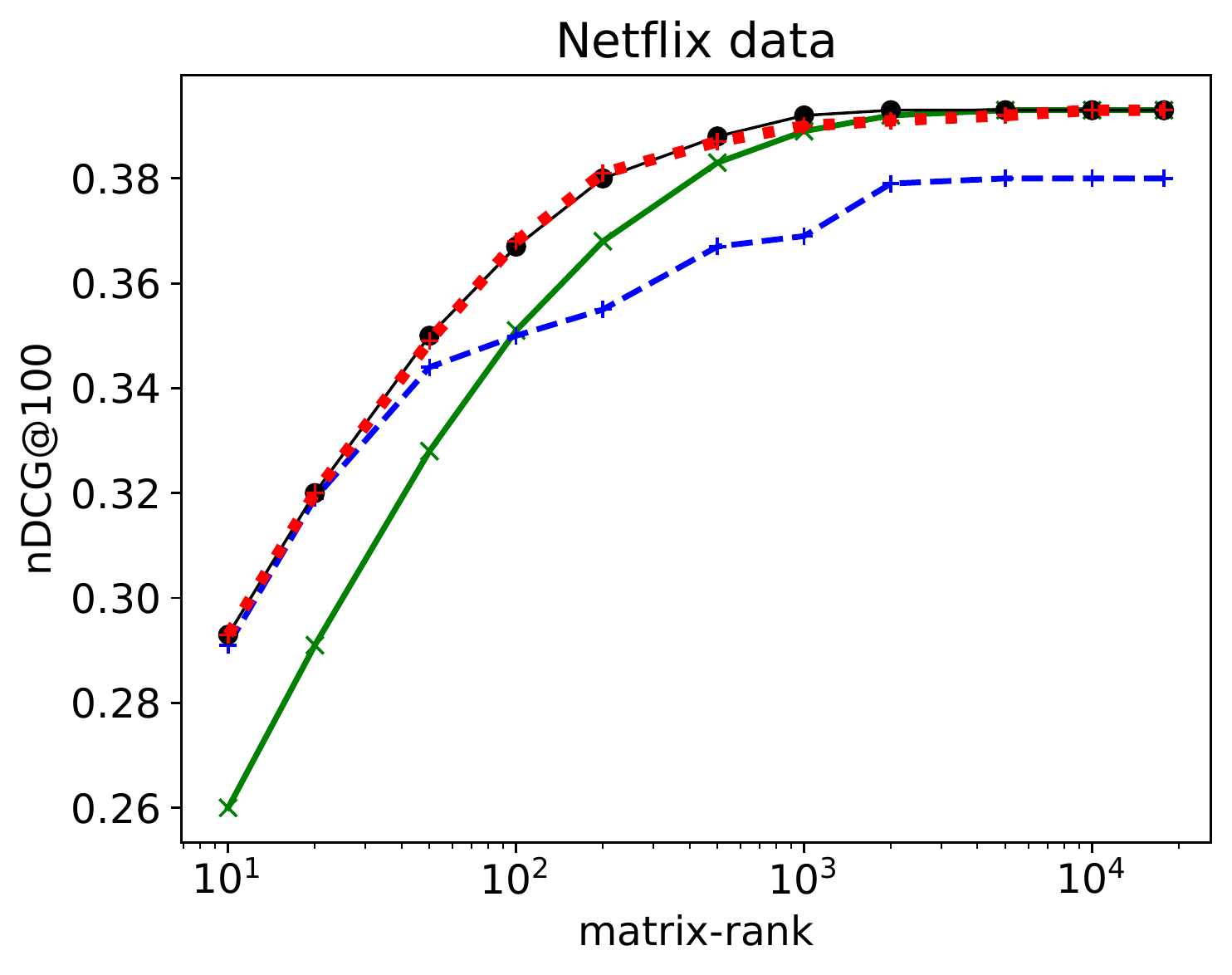}
 \includegraphics[width=58mm]{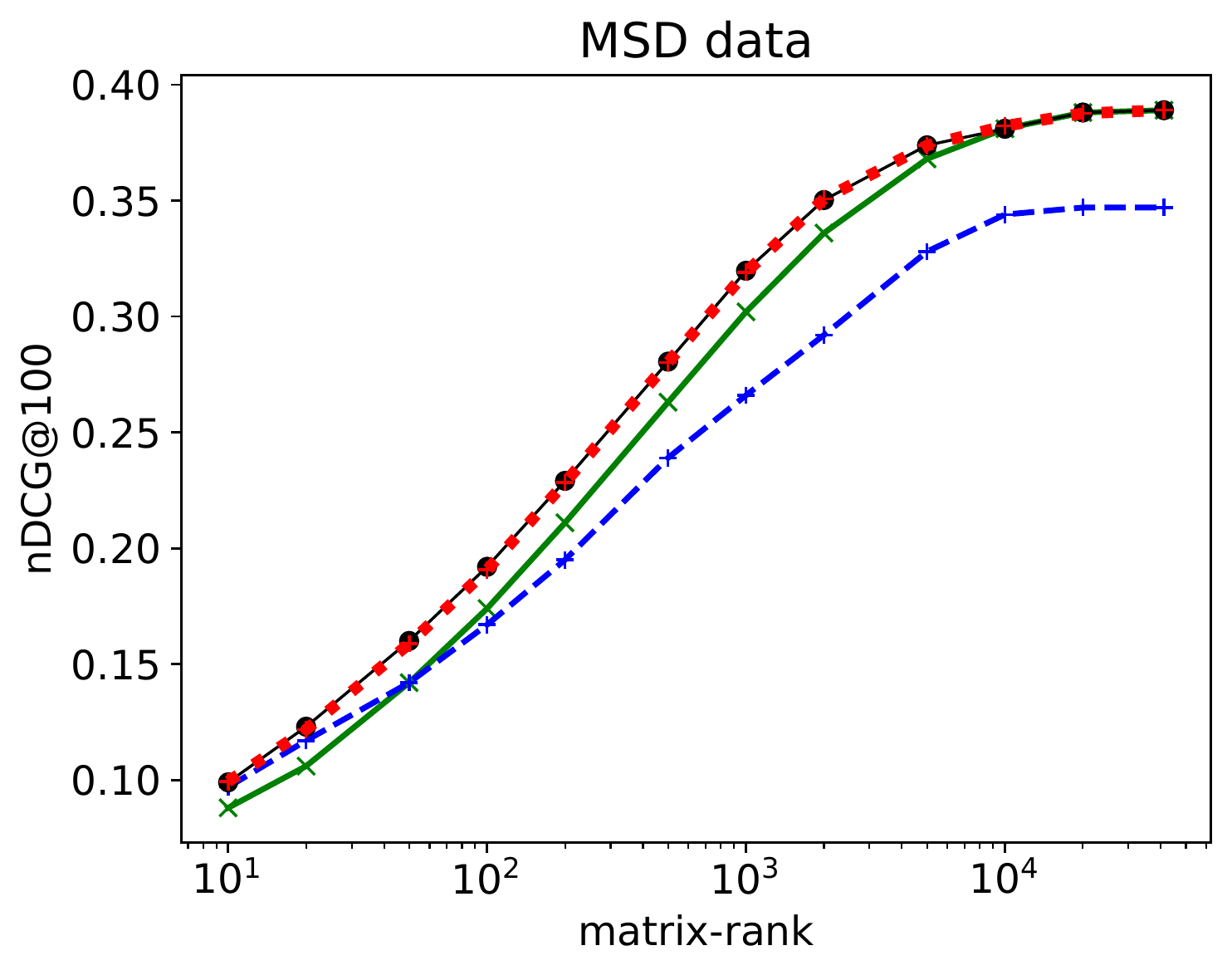}
 \caption{Ranking accuracy for different model-ranks $k={\rm rank}(\UU\VV^\top)$ on the three data-sets \emph{ML-20M}, \emph{Netflix} and \emph{MSD}, where the standard errors are 0.002, 0.001, and 0.001, respectively: the proposed closed-form solution (red dotted line) provides an accurate approximation to the exact solution of  EDLAE  (black line) across \emph{all} ranks $k$. Like in \cite{steck20b}, also the models with the constraint $\diag(\UU\VV^\top)=0$ (green solid line) and without any constraints on the diagonal (blue dashed line) are shown for comparison.}
 \label{fig_res}
\end{figure}

We reproduced figure 1 from \cite{steck20b} for the Netflix Prize data, which shows the ranking accuracy (nDCG@100) on the test data for various model-ranks $k={\rm rank}(\UU\VV^\top)$, ranging from 10 to full-rank. These results, obtained by iteratively optimizing the training objective using the update-equations of ADMM derived in \cite{steck20b},  
serve as the exact solution (ground truth) in our experiments.
In addition to the \emph{Netflix} data, we also generated the same kind of graph for the MovieLens 20 Million data (\emph{ML-20M}) as well as the Million Song Data (\emph{MSD}) using the publicly available code of \cite{steck20b}. We tuned the L$_2$-norm regularization $\Lambda$ for each of the different model-ranks $k$ as to optimize prediction accuracy of each model.

We then added the results based on the closed-form solution derived above (red dotted line in Figure \ref{fig_res}): as we can see in Figure \ref{fig_res}, the proposed closed-form solution (red dotted line) provides an excellent approxiamtion to the (exact) solution obtained by ADMM in \cite{steck20b} (black solid line)  \emph{across the entire range} of model-ranks $k$. This is particualrly remarkable, as the other two approximations considered in \cite{steck20b} (green solid line and blue dashed line in Figure \ref{fig_res})  only provide an accurate approximation for either very large or very small model-ranks $k$, respectively. The green solid line reflects the model with the constraint $\diag(\UU\VV^\top)=0$ applied to Eq. \ref{eq_trainuv_orig}, while the blue dashed line is based on the unconstrained linear AE, i.e., $\min_{\UU,\VV} \lVert \XX - \XX \UU\VV^\top \rVert_F^2 + \lVert \Lambda^{\nicefrac{1}{2}}\UU\VV^\top \rVert_F^2$.

These results also imply that, despite our 'initialization' $\beta=0$ for the diagonal values in our approach (see item 2 in Section \ref{sec_derivation}), the proposed closed-form solution is indeed able to learn the approximately correct values of the diagonal $\diag(\UUhat\VVhat^\top)$, which can be markedly different from zero at small model-ranks $k$.
This is  implied by the fact that the unconstrained AE, which obviously has a non-zero diagonal (blue dashed line), is close to the proposed approximation (red dotted line) in Figure \ref{fig_res} for small $k$. On the other hand, at large model-ranks $k$,  the proposed closed-form solution learns that  $\diag(\UUhat\VVhat^\top)$ is indeed close to zero, as implied by the fact that the proposed approximation (red dotted line) is close to the model trained with the constraint  $\diag(\UU\VV^\top)=0$   (green solid line) in Figure \ref{fig_res} for large $k$.

Apart from that, Figure \ref{fig_res} also shows that the proposed approximation of EDLAE (red dotted line in Figure \ref{fig_res}) achieves a considerable gain in ranking-accuracy on test-data, compared to the exact closed-form solution of the unconstrained linear AE (blue dashed line), which was  derived in \cite{jin21}.

\section*{Conclusions}

In this paper, we used simple properties of the singular value decomposition (SVD) to gain insights into the training of autoencoders in two scenarios.

First, we showed that  \emph{unsupervised} training by itself induces  severe regularization / reduction in model-capacity of \emph{deep nonlinear} autoencoders. This is  reflected by the fact that \emph{deep nonlinear} autoencoders \emph{cannot} fit the data more accurately than \emph{linear} autoencoders do if they have the same dimensionality in their last hidden layer (and under a few additional mild assumptions). We discussed several aspects of this new and counterintuitive insight, including that this might provide the first explanation for the puzzling experimental observations in the literature of recommender systems that deep nonlinear autoencoders struggled to achieve markedly higher ranking-accuracies than simple linear autoencoders did.

Second, we derived a simple teacher-student algorithm for approximately learning a low-rank EDLAE model \cite{steck20b}, a linear autoencoder that is regularized to not overfit towards the identity-function. We empirically observed that the derived closed-form solution provides an accurate approximation across the entire range of latent dimensions. Moreover, except for very small numbers of latent dimensions, it considerably outperformed the exact closed-form solution of the unconstrained linear autoencoder  derived in \cite{jin21}.

\subsubsection*{Acknowledgements}
We are grateful to useful comments from and discussions with Dawen Liang, Nikos Vlassis, Ehtsham Elahi,  and Justin Basilico.

\section*{Appendix: Proof of Proposition}
In this proof of the Proposition in Section \ref{sec_prop}, two mathematically simple facts are combined. The singular value decomposition (SVD) plays a pivotal role, which we relate to the deep nonlinear AE in the first step, and in the second step to the linear AE. The 
least-squares problem
\begin{equation}
 \min_{M: {\rm rank}(M)\le k}|| X- M ||_F^2 ,
 \label{eq_m}
\end{equation}
where $M \in \RR^{m \times n}$ is a matrix with a rank of (at most) $k<\min(m,n)$, has a well-known solution (Eckart–Young–Mirsky theorem): let
\begin{equation}
 X= U\cdot S \cdot V^\top
 \label{eq_svd}
\end{equation}
denote the SVD of $X$, where $U \in \RR^{m\times \min(m,n)}$ and $V \in \RR^{n \times \min(m,n)}$ are the matrices with the left and right singular vectors, respectively, while $ S \in \RR^{\min(m,n)\times \min(m,n)}$ is the diagonal matrix of the singular values of $X$. Then the rank-$k$ solution that minimizes the squared error in Eq. \ref{eq_m} is 
\begin{equation}
 M^*= U_k\cdot S_k \cdot V_k^\top ,
 \label{eq_m_solution}
\end{equation}
where $ S_k \in \RR^{k \times k}$ is the diagonal matrix with the $k$ largest singular values in $S$;  $U_k \in \RR^{m\times k}$ and $V_k \in \RR^{n \times k}$ are the sub-matrices of $U$ and $V$ with the singular vectors corresponding to these $k$ largest singular values.

Now, in the first step, we can make the connection to the \emph{deep nonlinear} AE: when comparing Eq. \ref{eq_m_solution} to Eq. \ref{eq_deep}, note that $g_{\theta'}(X)$ in the nonlinear AE may be viewed as a parametric model, while the corresponding matrix $U_k$ in $M^*$ may be viewed as a non-parametric approach.\footnote{'Non-parametric' is used here in the sense that, for each data-point (i.e., row) in the given training-data matrix $X$, there is a corresponding row (with $k$ model-parameters) in matrix $U_k$, i.e., the number of model parameters in $U_k$ grows linearly with the number of data-points in $X$. In contrast, $g_{\theta'}(X)$ in the nonlinear AE is a parametric model, as the function $g_{\theta'}(\cdot)$ has a fixed number of parameters, and it is applied to each row of $X$.}
Obviously, the parametric model cannot fit the training-data more accurately than the non-parametric approach does. On the other hand, $W_L$ in the nonlinear AE and the corresponding $S_k \cdot V_k^\top$ in $M^*$ are both (non-parametric) matrices of rank $k$. This implies that the deep nonlinear AE in Eq. \ref{eq_deep}  cannot fit the training data better than $M^*$ in Eq. \ref{eq_m_solution} does. Hence, we have 
\begin{eqnarray}
 {\rm SE}^{\rm(deep)} \!\!\!\!\!\! &=& \!\!\!\!\!\! \min_{\theta', W_L}|| X- g_{\theta'}(X) \cdot W_L ||_F^2\nonumber \\
 &\ge& \!\!\!\!\!\! ||X - U_k\cdot S_k \cdot V_k^\top||_F^2 = || X- M^* ||_F^2 .
 \label{eq_proof_deep}
\end{eqnarray}
Now, in the second step, we can see that $M^*$ (non-parametric approach) is actually equivalent to  the \emph{linear} AE (parametric model), due to the simple identity
\begin{equation}
U_k\cdot S_k \cdot V_k^\top = U\cdot S \cdot \underbrace{V^\top \cdot V_k}_{=I_{n\times k}} \cdot V_k^\top ,
\end{equation}
which follows immediately from the fact that $I_{n\times k} \in \RR^{n\times k}$ is essentially a (rectangular) identity-matrix of rank $k$, which then selects the submatrices $S_k$ from $S$, and $U_k$ from $U$. This simple equation seems to have been overlooked until it was pointed out in \cite{jin21}. With Eqs. \ref{eq_m_solution} and \ref{eq_svd}, it now follows \cite{jin21}
\begin{equation}
M^*= U_k\cdot S_k \cdot V_k^\top = U\cdot S \cdot V^\top \cdot (V_k \cdot V_k^\top) = X \cdot (V_k \cdot V_k^\top).
\end{equation}
Comparing this to the linear AE in Eq. \ref{eq_lin}, we can see that one of possibly many solutions of Eq. \ref{eq_se_lin} is $W_1 := W_2^\top := V_k$. Hence,
the rank-$k$ SVD (non-parametric approach) and the linear AE (parametric model) can fit the training data equally well:
\begin{eqnarray}
 {\rm SE}^{\rm(linear)} \!\!\!\!\!\! &=& \!\!\!\!\!\! \min_{W_1,W_2}|| X- X \cdot W_1 \cdot W_2 ||_F^2 \nonumber\\
 &=& \!\!\!\!\!\! ||X - X \cdot V_k \cdot V_k^\top||_F^2 = || X- M^* ||_F^2
 \label{eq_proof_lin}
\end{eqnarray}
Comparing Eqs. \ref{eq_proof_deep} and \ref{eq_proof_lin}, we obtain the claim in Eq. \ref{eq_prop}, i.e., ${\rm SE}^{\rm(deep)} \ge {\rm SE}^{\rm(linear)}$. $\Box$
  
\bibliographystyle{apalike}
\bibliography{}

\end{document}